\documentclass[journal]{IEEEtran}

\usepackage{comment}
\usepackage{epsfig}
\usepackage{graphicx}
\usepackage{amsmath}
\usepackage{amssymb}
\usepackage{comment}
\usepackage{multirow}

\usepackage{hyperref}
\hypersetup{hidelinks}

\usepackage{booktabs}
\usepackage{array, caption, threeparttable}
\usepackage{caption}
\usepackage{subfigure}
\usepackage[ruled]{algorithm2e}
\usepackage{listings}
\usepackage{color}
\usepackage{mathtools}

\usepackage{todonotes}
\usepackage{gensymb}
\usepackage{setspace}
\usepackage[normalem]{ulem}
\usepackage{cancel}

\begin{document}

\title{Semi-rPPG: Semi-Supervised Remote Physiological Measurement with Curriculum Pseudo-Labeling}

\author{Bingjie Wu, Zitong Yu, Yiping Xie, Wei Liu, Chaoqi Luo, Yong Liu, Rick Siow Mong Goh

\thanks{Manuscript received August, 2024. This work was supported by National Natural Science Foundation of China No. 62306061. Corresponding author: Zitong Yu (email: zitong.yu@ieee.org)}

\thanks{B. Wu, Y. Liu, W. Liu and S. Gohare are with Institute of High Performance Computing (IHPC), Agency for Science, Technology and Research (A*STAR), Singapore.}

\thanks{Z. Yu and C. Luo are with School of Computing and Information Technology, Great Bay University, Dongguan, 523000, China.}

\thanks{Y. Xie is with Computer Vision Institute, School of Computer Science \& Software Engineering, Shenzhen University, Shenzhen, 518060, and also with School of Computing and Information Technology, Great Bay University, Dongguan, 523000, China.}

}

\markboth{IEEE Transactions on Instrumentation and Measurement}%
{Shell \MakeLowercase{\textit{et al.}}: Bare Demo of IEEEtran.cls for IEEE Journals}

\maketitle


\begin{abstract}

Remote Photoplethysmography (rPPG) is a promising technique to monitor physiological signals such as heart rate from facial videos. However, the labeled facial videos in this research are challenging to collect. Current rPPG research is mainly based on several small public datasets collected in simple environments, which limits the generalization and scale of the AI models. Semi-supervised methods that leverage a small amount of labeled data and abundant unlabeled data can fill this gap for rPPG learning. In this study, a novel semi-supervised learning method named Semi-rPPG that combines curriculum pseudo-labeling and consistency regularization is proposed to extract intrinsic physiological features from unlabelled data without impairing the model from noises. Specifically, a curriculum pseudo-labeling strategy with signal-to-noise ratio (SNR) criteria is proposed to annotate the unlabelled data while adaptively filtering out the low-quality unlabelled data. Besides, a novel consistency regularization term for quasi-periodic signals is proposed through weak and strong augmented clips. To benefit the research on semi-supervised rPPG measurement, we establish a novel semi-supervised benchmark for rPPG learning through intra-dataset and cross-dataset evaluation on four public datasets. The proposed Semi-rPPG method achieves the best results compared with three classical semi-supervised methods under different protocols. Ablation studies are conducted to prove the effectiveness of the proposed methods.


\end{abstract}


\IEEEpeerreviewmaketitle

\section{Introduction}

\IEEEPARstart{R}{emote} Photoplethysmography (rPPG) \cite{yu2021facial} is a non-invasive technique to monitor blood volume changes in the microvascular tissue of the skin. Numerous research studies have been conducted to utilize rPPG technique to measure physiological signals including heart rate \cite{tran2015robust,hu2021robust,yue2021deep,das2023time,hu2021eta}, blood pressure \cite{wu2022facial,schrumpf2021assessment,wu2022camera}, oxygen saturation \cite{zhu2022contactless}, respiratory rate \cite{du2021weakly,alnaggar2023video}, and diseases such as Atrial Fibrillation \cite{shi2019atrial}. In contrast to traditional measurement devices such as oximeters and blood pressure monitors, the rPPG measurement is non-contact, convenient, and cost-effective, making it a promising technique in the digital health field.

\begin{figure}
\centering


\subfigure[Supervised methods]{
		\begin{minipage}[b]{0.5\textwidth}
			\includegraphics[width=1\textwidth]{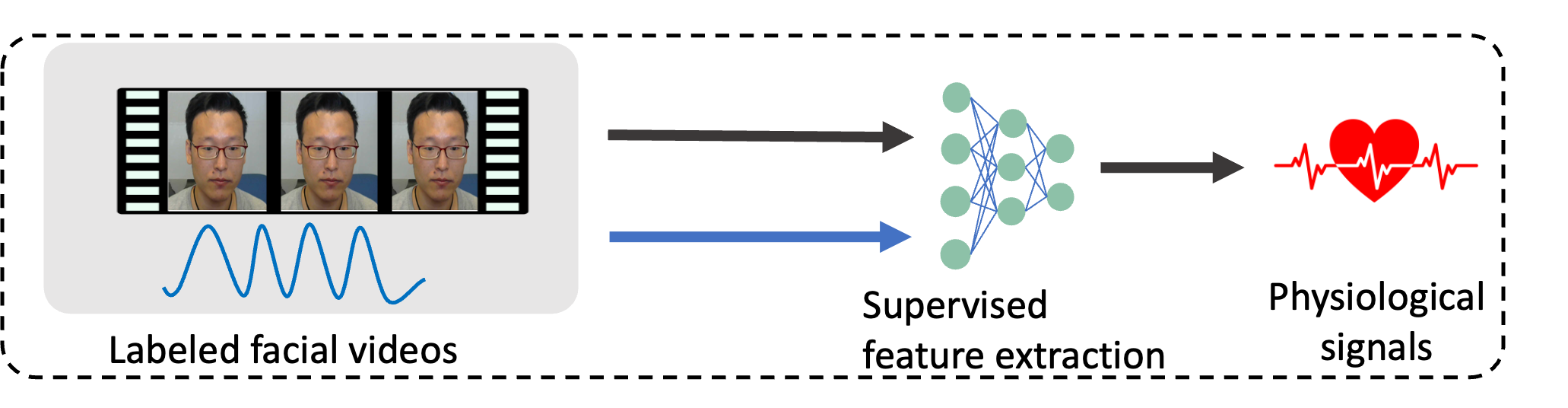}
		\end{minipage}
		\label{fig:sup}
	}\vspace{-5mm}
 \subfigure[Self-supervised methods]{
		\begin{minipage}[b]{0.5\textwidth}
			\includegraphics[width=1\textwidth]{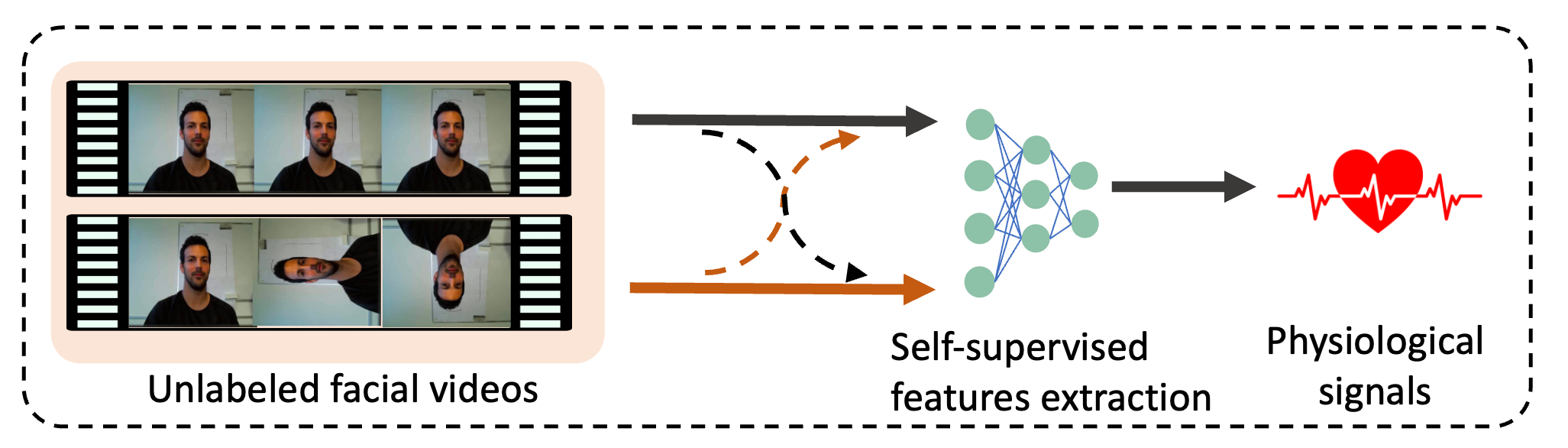}
		\end{minipage}
		\label{fig:selfsup}
	}\vspace{-5mm}
  \subfigure[Semi-supervised methods]{
		\begin{minipage}[b]{0.5\textwidth}
			\includegraphics[width=1\textwidth]{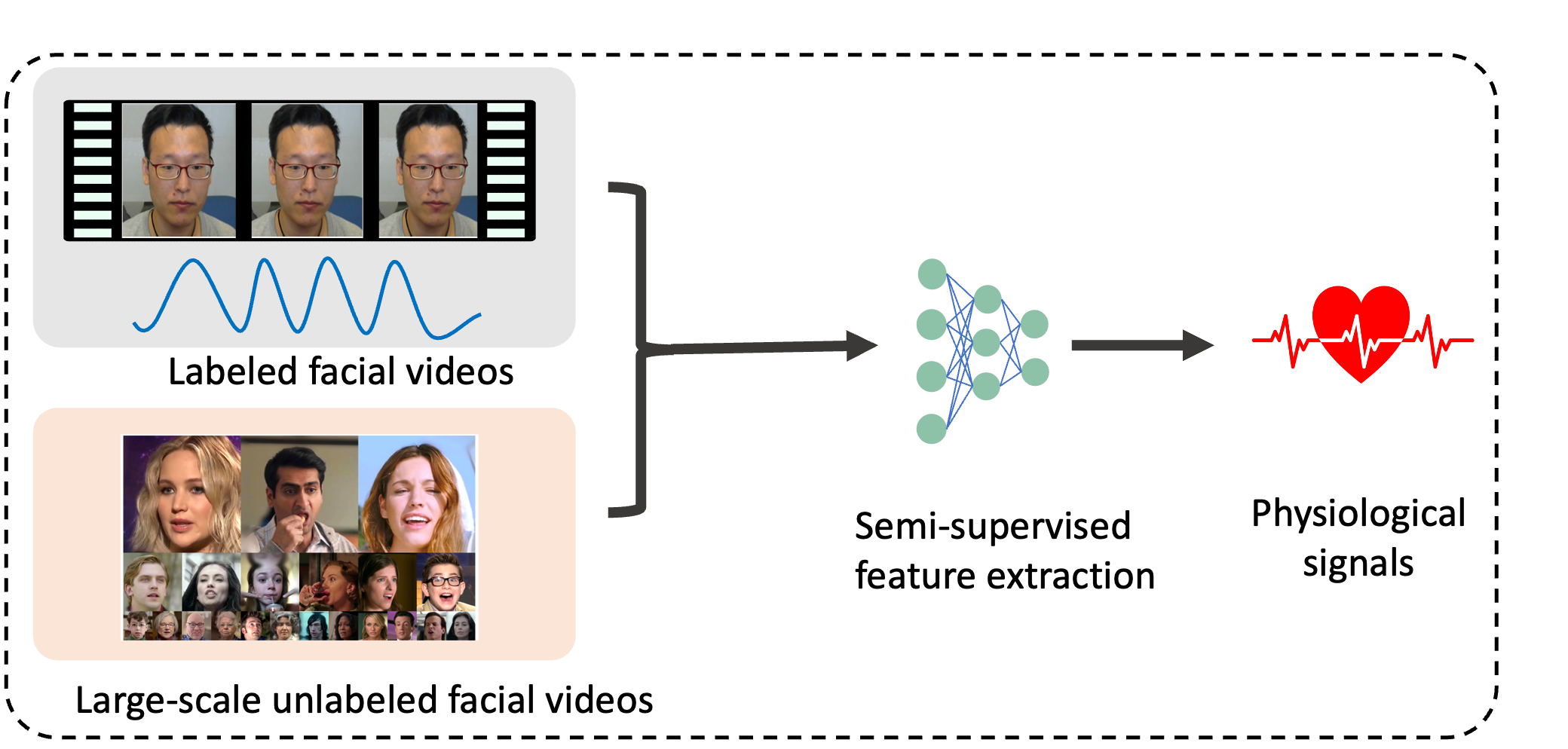}
		\end{minipage}
		\label{fig:semisup}
	}\vspace{-3mm}
\caption{Comparison of deep learning methods for rPPG learning: (a) Supervised methods: utilize labeled data for training; (b) Self-supervised methods: leverage unlabeled data for training; (c) Semi-supervised methods: use a small amount of labeled data and a large amount of unlabeled data for training}
\label{fig:compare}
\end{figure}

The current research on rPPG deep learning concentrates on supervised methods and self-supervised methods as indicated in Fig.~\ref{fig:sup} and Fig.~\ref{fig:selfsup}. The supervised method is to use the labeled facial videos to extract the rPPG signal. The self-supervised method is to learn the rPPG signal from unlabeled facial videos through its characteristics. The semi-supervised method of rPPG learning (Fig.~\ref{fig:semisup}) aims to leverage small labeled data and abundant unlabeled data. It has not been sufficiently studied. The commonly used public facial video-based rPPG datasets are VIPL-HR~\cite{niu2019vipl}, PURE~\cite{stricker2014non}, UBFC-rPPG~\cite{bobbia2019unsupervised}, COHFACE~\cite{heusch2017reproducible}, and so on. The numbers of subjects for these datasets are 107,10,42, and 40, respectively. The colored videos for each dataset are 2378, 60, 42, and 160, respectively. The number of subjects and videos is quite limited. The labeled dataset of rPPG research consisting of facial videos with concurrent rPPG signals is very difficult to collect as privacy data. Current research is based on several small datasets collected under simple environments, which limits the generability and scale of the deep learning models. On the other hand, there are abundant available facial videos without ground-truth labels such as the CelebV-HQ dataset \cite{zhu2022celebv}. Semi-supervised methods can leverage both labeled and unlabeled data which can further enhance the rPPG learning. 
 
\textbf{}To cover the above gaps, a novel semi-supervised learning method of Semi-rPPG is proposed to leverage small labeled facial videos and abundant unlabeled facial videos in this study. A curriculum pseudo-labeling strategy and a consistency loss for quasi-periodic signals are proposed to reduce the disturbance of noisy labels. Firstly, the model is trained with the labeled data with supervised losses and consistency loss in the first epoch. Secondly, pseudo-labels are generated for the unlabeled data with the curriculum pseudo-labeling strategy. Next, the labeled data and pseudo-labeled data will be merged for supervised training.  

Our main contributions to this article are as follows:

    
    
    

\begin{enumerate} 
  \item To address the gap of insufficient semi-supervised rPPG studies, we propose a novel semi-supervised learning method, Semi-rPPG, which enhances the model performance by leveraging abundant unlabelled data. 
  \item We design a curriculum pseudo-labeling strategy that can adaptively employ high-quality pseudo-labels for model training while mitigating the impact of noisy pseudo-labels.
  \item We introduce a novel consistency loss that leverages weak and strong augmented clips to extract specific temporal features from quasi-periodic signals.
  \item To the best of our knowledge, this study is the first to establish a semi-supervised rPPG benchmark using both intra-dataset and cross-dataset settings. We implement and compare our approach with three traditional semi-supervised methods, and conduct extensive ablation studies to thoroughly evaluate its performance.

\end{enumerate}

In the rest of the paper, Sec.~\ref{sec:relatedwork} provides the related work of rPPG learning. Sec.~\ref{sec:method} formulates the Semi-rPPG structure, and introduces the curriculum pseudo-labeling and consistency regularization designed for quasi-periodic signals. Sec.~\ref{sec:experiment} presents the implementation details and evaluates the performance of the proposed method on four benchmark datasets examining both intra-dataset and cross-dataset scenarios. Ablation studies are conducted and the visualization results are plotted in this section. Finally, a conclusion with future directions is given in Sec.~\ref{sec:conclusion}.

\begin{figure*}
\centering
\includegraphics[scale=0.7]{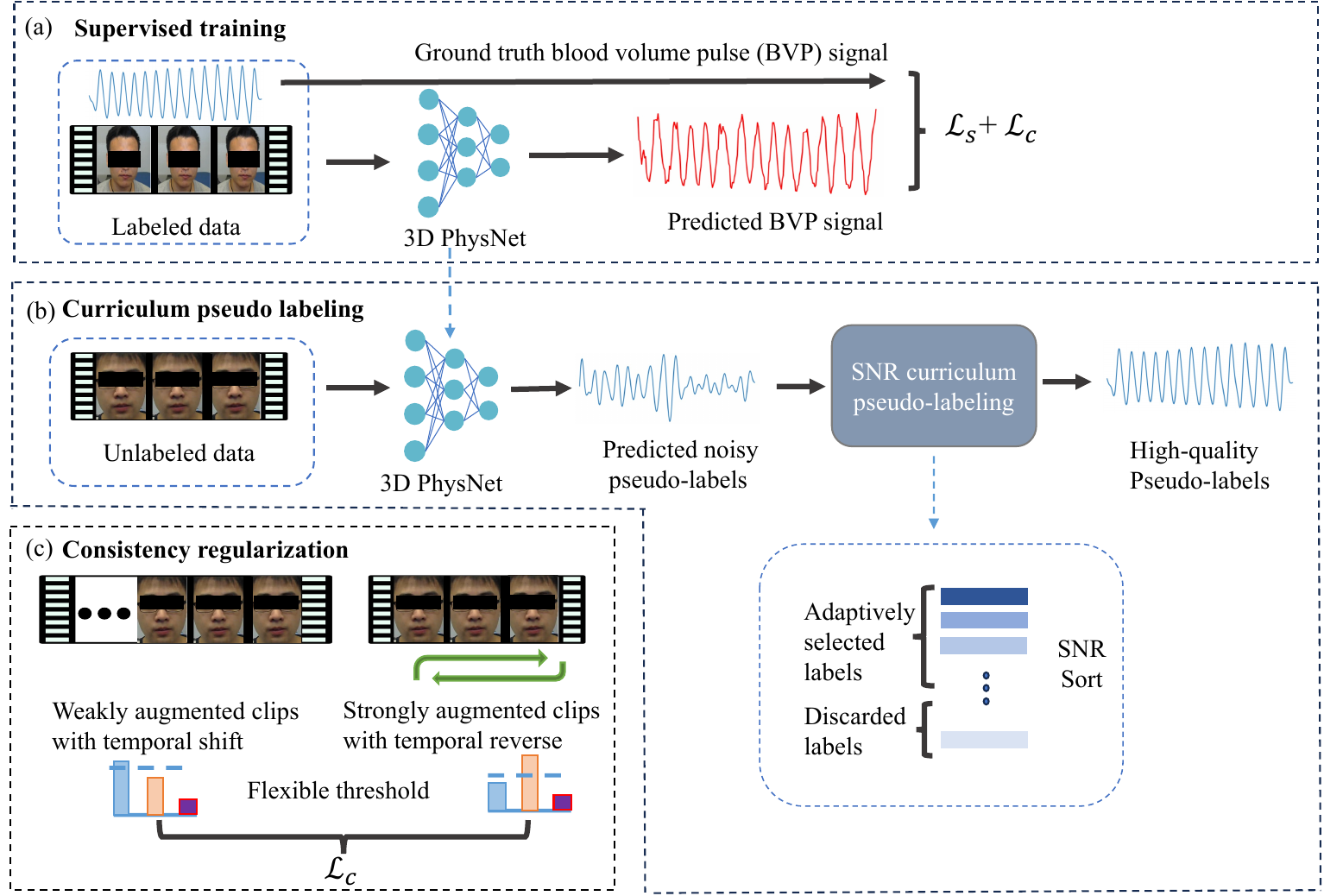}
  \caption{\small{
  The overall framework of the proposed Semi-rPPG. (a) In the supervised stage, the model is trained on labeled data using supervised loss and consistency regularization. (b) During the pseudo-labeling stage, the trained model assigns preliminary pseudo-labels to the unlabeled data. Next, the signal-to-noise ratio (SNR) is computed and sorted for all unlabeled samples. Top high-quality pseudo-labels are then adaptively chosen based on a curriculum ratio $R$. These selected pseudo-labels, along with the corresponding unlabeled data, are combined with the labeled dataset and used for iterative supervised training, as illustrated in section (a). (c) The facial clip is weakly and strongly augmented with a small temporal shift and a temporal reverse, respectively. A consistency regularization term is proposed as the heart rate shall remain consistent across these two augmentations.
  }
  }
  
\label{fig:overview}
\end{figure*}

\section{Related Work}
\label{sec:relatedwork}

\subsection{Traditional rPPG measurement}
As the rPPG signal change is very subtle, it is difficult to directly achieve from videos. Some traditional methods such as CHROM \cite{de2013robust} and POS~\cite{wang2016algorithmic} adopted mathematical models to decompose the videos into rPPG-relevant parts and non-relevant parts. However, the environment of illumination conditions and motions and the individual skin conditions are complicated to be incorporated into mathematical models. Besides, the input video is 3-dimensional (3D) signals involving spatial and temporal information which adds difficulty in developing the mathematical models. The deep-learning methods, however, are capable of extracting the rPPG signal from complex environments and individual conditions. 

\subsection{rPPG deep learning methods}
The primary deep learning methods for rPPG are focused on supervised methods \cite{dong2024realistic,du2023dual,yu2019remote1,yu2023physformer++} and self-supervised methods \cite{wang2022self,sun2022contrast,gideon2021way,speth2023non,liu2024rppg,park2022self}. Previous supervised methods \cite{niu2019rhythmnet} used neural networks to regress heart rate as a single value. However, the characteristics of rPPG signals as quasi-periodic waveforms were neglected and much useful information was not leveraged in training models. The recent supervised methods predict the rPPG signal to leverage characteristics of signals by adding constraints of time and frequency domains in terms of signal similarity \cite{niu2020video}, signal distribution \cite{yu2022physformer}, and power spectral density (PSD) distance~\cite{gideon2021way}.

For self-supervised studies, some research utilizes the temporal and spatial characteristics of facial videos to construct self-supervised loss.  A contrastive framework was proposed by Su and Li \cite{sun2022contrast}. The positive samples were obtained from the anchor sample's different facial areas and near temporal windows. The negative samples were selected from different individuals' face videos. However, there are possibilities that different individuals have similar PSD which will confuse the model to push the negative samples. Since the rPPG signal is quasi-periodic, many self-supervised methods are developed taking advantage of its frequency characteristics. Jeremy Speth et al.~\cite{speth2023non} proposed a non-contrastive unsupervised learning framework. The model was constrained by the bandlimits, variance, and sparsity of the power spectral of the rPPG signal. The proposed framework was valid under a stable and simple environment. The authors tested that the model was not converged with a noisy CelebV-HQ dataset \cite{zhu2022celebv}.

As the input of the rPPG learning is 3D video clips,  some research \cite{niu2019rhythmnet,niu2020video,das2021bvpnet} pre-process the input into 2D spatial-temporal maps to adopt the off-the-shelf 2D models by averaging the pixel values of several region of interest (ROI) areas to increase the robustness of the model.  While the other research uses end-to-end 3D neural networks~\cite{yu2019remote,yu2022physformer} without pre-processing the input facial clips. 


\subsection{Semi-supervised rPPG learning for vision tasks}
Semi-supervised learning methods have been widely studied in image classification \cite{tarvainen2017mean}, object detection \cite{li2022rethinking}, semantic segmentation \cite{alonso2021semi}, and action recognition tasks \cite{jing2021videossl}. The major semi-supervised methods can be roughly divided into consistency regularization and pseudo-labeling \cite{he2023pseudo}. The consistency regularization assumes the model will have the same output under spatiotemporal similarity or data augmentation. While pseudo-labeling annotates pseudo-labels for the unlabeled data. However, the pseudo-labels may also underperform due to noises. FixMatch \cite{sohn2020fixmatch} is a traditional SSL method assuming the consistency between weekly augmented images and strongly augmented images. A fixed threshold is adopted on the predicted class probability to filter out high-quality unlabeled data for training. The drawback of FixMatch is that it assumes an equal threshold for all classes without considering the learning difficulties between different classes. To improve FixMatch, FlexMatch \cite{zhang2021flexmatch} was proposed by replacing the fixed threshold with a flexible threshold. This method will calculate a dynamic threshold for each class at each time step. Hasan et al.~\cite{hasan2023srppg} proposed a semi-supervised adversarial learning framework that utilized two temporal consistency properties. It assumes two partially overlapped video frame sequences shall contain the same PPG signals for the overlapped period. Besides, the downsampled video frames will correspond to the downsampled PPG signal. According to the authors' statistics, research on semi-supervised rPPG learning is not sufficient. This study aims to address that gap by exploring rPPG learning using semi-supervised methods.


\begin{figure*}[t]
\centering
\includegraphics[scale=0.6]{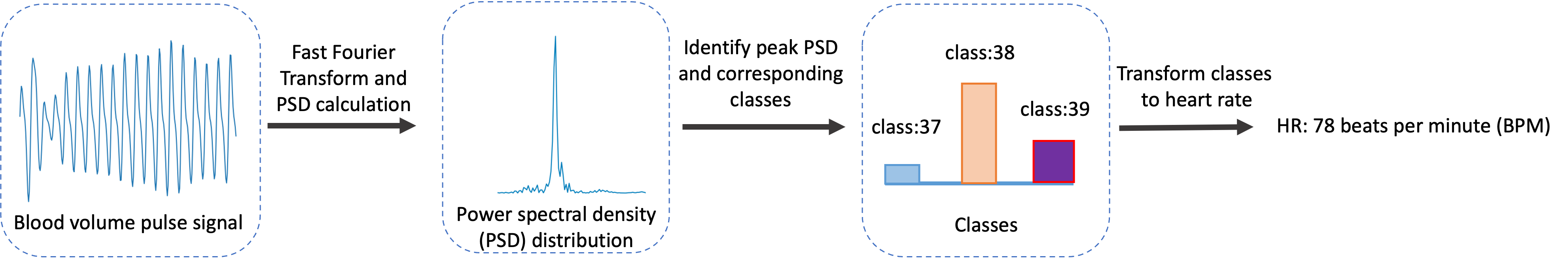}
  \caption{\small{
  An illustration of calculating heart rate from blood volume pulse signal (BVP): First, the BVP signal is converted into power spectral density (PSD) through Fast Fourier Transform; Next, the PSD is categorized into classes ranging from 0 to 140, which correspond to heart rates of 40 to 180  beat per minute (BPM); Finally, the heart rate is determined by reverse mapping the class.
  }
  }
\label{fig:bvp_to_classes}
\end{figure*}

\section{Methodology}
\label{sec:method}

In this study, we propose a semi-supervised approach for extracting rPPG signals from facial videos, referred to as the Semi-rPPG. Blood volume pulse (BVP), as a common rPPG signal, is used to represent the rPPG data in this study. An overview of the proposed method is depicted in Fig.~\ref{fig:overview}.  The task is to map the face video clips $X \in \mathbb{R}^{T\times W \times H\times C}$ to the BVP signals $Y \in \mathbb{R}^{T}$, where $T,W,H,C$ represent the frame, width, height, and channel of the videos. An end-to-end 3D-CNN of PhysNet \cite{sun2022contrast} $f_{\theta}:X\rightarrow{Y}$ is utilized to capture the spatial-temporal features in this study. First, the labeled data of facial clips, which includes synchronized BVP signals, are used to train the model for the first epoch. The model then generates the pseudo BVP labels for the unlabelled facial videos. Following a curriculum noise depression strategy, only top high-quality pseudo-labels are employed for the next epoch's training. Starting from the second epoch, the labeled facial clips are combined with the unlabeled facial clips with pseudo-labels, creating a unified dataset for iterative training. Besides, a self-supervised consistency loss leveraging weak and strong augmented clips is proposed to extract specific temporal features from quasi-periodic signals. All the labeled and unlabeled clips are employed in the calculation of this consistency loss. The detailed semi-rPPG algorithm is demonstrated in algorithm~\ref{algorithm:semi-rppg}.



\begin{algorithm}[t]
\caption{Semi-rPPG Method}
\label{algorithm:semi-rppg}
\LinesNumbered
\small
\KwIn {Labeled video clips \{$X_{l},Y_{l}$\} and unlabeled video clips \{$X_{un}$\}}
\KwOut{$\theta_{f}$ }
\For {epoch $e=0$} { 
Sample labeled clips $x_{l},y_{l}$\;
Generate weakly and strongly augmented clips $x_{w},x_{s}$\;
Calculate supervised loss and consistency loss: $ L_{0}=L_{s}(x_{l},y_{l})+L_{c}(x_{w},x_{s})$\;
Update parameters $\theta_{f}$ by descending $\nabla_{f}L_{0}$\;
Generate pseudo-labels through Eq.(\ref{eq:caly})\;
Sort and filter out high-quality pseudo-labels $Y_{sel}$ and corresponding clips $X_{sel}$ through Eq.(5-8)\;
Merge dataset of \{$X_{l},Y_{l}$\} with \{$X_{sel},Y_{sel}$\} into a new dataset\{$X_{new},Y_{new}$\}\;
}

\For {epoch $e=1:e_{total}$} {
Sample clips $x_{l},y_{l}$ from \{$X_{new},Y_{new}$\} and  ${x}_{un}$ from \{$X_{un}$\} \;
Generate weakly and strongly augmented clips $x_{w\_l}$, $x_{s\_l}$, $x_{w\_un}$, $x_{s\_un}$ \;
Calculate supervised loss and consistency loss $ L=L_{s}(x_{l},y_{l})+L_{c}(x_{w\_l},x_{s\_l})+L_{c}(x_{w\_un},x_{s\_un})$ \;
Update architecture $\theta_{f}$ by descending $\nabla_{f}L$  \;
Generate pseudo-labels through Eq.(\ref{eq:caly})\;
Sort and filter out high-quality pseudo-labels $Y_{sel}$ and corresponding clips $X_{sel}$ through Eq.(5-8)  \;
Merge dataset of \{$X_{l},Y_{l}$\} with \{$X_{sel},Y_{sel}$\} into a new dataset \{$X_{new},Y_{new}$\}\;
}
\end{algorithm}

\subsection{Supervised loss}
\label{sec:suploss}

As the heart rate is normally an integer because it counts discrete events of contractions of the heart per minute and the ground truth of heart rate provided by the benchmark dataset of VIPL-HR-V1 are all integers ranging from 40-180 beats per minute (BPM), the heart rate prediction can be considered as a classification problem. This study follows the protocol of this paper~\cite{yu2022physformer} converting the regression loss of pulse signal prediction as the classification loss. The heart rates of 40-180 BPM are classified as classes 0-140. For example, the heart rate of 40 BPM corresponds with class 0 and the heart rate of 41 BPM corresponds with class 1. An illustration of heart rate calculation from BVP signal is depicted in Fig.~\ref{fig:bvp_to_classes}. First, a sequence of BVP signals is converted from the time domain to the frequency domain using the Fast Fourier Transform. Frequencies outside the range of 0.67-3 Hz are deemed as noise and removed. The power spectral density (PSD) distributions for the sequence are then computed, and the peak PSD is identified. The heart rate class is determined based on the frequency associated with the peak PSD.

The supervised loss $L_{s}$ consists of the cross entropy loss of $L_{ce}$ and the negative Pearson correlation loss $L_{p}$. The $L_{p}$ is adopted to evaluate the linear correlation of the predicted blood volume pulse $\hat{y}$ with the ground truth signals $y$.


\begin{equation}
\begin{aligned}
L_{ce}=CE(PSD(\hat{y}), \mathbf{e}(HR)) 
\end{aligned}
\label{eq:ce}
\end{equation}
where $CE$ is the cross-entropy loss, $\hat{y}$ is the predicted BVP, $PSD(\hat{y})$ represents the power spectral density of $\hat{y}$, and $\mathbf{e}(HR)$ denotes the one-hot encoding of true heart rate class $HR$.


\begin{equation}
\begin{aligned}
L_{p}=1-r
\end{aligned}
\label{eq:pearson}
\end{equation}

\begin{equation}
\begin{aligned}
r=\frac{\sum_{i=1}^{T}(\hat{y}_{i}-\hat{\overline{y}})(y_{i}-\overline{y})}{\sqrt{\sum_{i=1}^{T}(\hat{y}_{i}-\hat{\overline{y}})^2}\sqrt{\sum_{i=1}^{T}(y_{i}-\overline{y})^2}}
\end{aligned}
\label{eq:negpearson}
\end{equation}
where $r$ is Pearson correlation coefficient, the $y$ is the ground-truth BVP, $\hat{y}$ is predicted BVP, $\hat{\overline{y}}$ is the average value of the predicted BVP, $\overline{y}$ is average value of ground-truth BVP, $T$ is the length of the BVP signal, and $y_i$ and $\hat{y}_i$ are the $i$-th element of $y$ and $\hat{y}$, respectively.

The supervised loss $L_{s}$ is:

\begin{equation}
\begin{aligned}
L_{s}=L_{ce}+\lambda L_{p}
\end{aligned}
\label{eq:sup}
\end{equation}
where $\lambda$ is a hyperparameter to trade off the contributions of the loss $L_{ce}$ and loss $L_{p}$.

\subsection{Curriculum pseudo-labeling}
\label{sec:currloss}
From the second epoch, the pseudo-labels are iteratively generated for unlabeled data according to Eq.(\ref{eq:caly}). Each unlabeled data has a corresponding pseudo-label. Given the limited amount of labeled data, the model is prone to bias in the early stage resulting in incorrect or noisy pseudo-labels. To address this, a dynamic selection strategy is designed to select high-quality pseudo-labels and reduce the impact of noise on the model as indicated in Eq.(\ref{eq:rank})-Eq.(\ref{eq:ratio}). The pseudo-label selection is based on two factors of ratio $R$ and signal-to-noise ratio (SNR) of the pseudo-labels as indicated in Eq.(\ref{eq:rank}). The pseudo-label whose SNR ranking top $k$ among all the pseudo-labels will be selected. $k$ is the number of the selected unlabeled data which is determined by the multiplication of ratio $R$ and $N_{un}$. $R$ denotes the ratio of the selected unlabeled data $X_{sel}$ among the entire unlabeled data. The R is a dynamic value based on the training epochs as calculated from the Eq.(\ref{eq:ratio}). SNR is employed to rank the pseudo-labels.  From the second epoch, the SNRs for predicted pseudo-labels are calculated based on Eq.(\ref{eq:SNR}) and ranked at each epoch. The pseudo-labels with higher SNRs represent high-quality labels with less noise. The top $R$ portion of ranked pseudo-labels and the corresponding clips are selected iteratively for the training in subsequent epochs. In the study, the ratio R is set at 20\% at the initial epoch and the 80\% at the last epoch. It is designed based on the rationale that as training progresses, the model becomes increasingly capable of generating high-quality pseudo-labels. The increase of $R$ allows a larger proportion of pseudo-labels to be selected with the model training.

\begin{equation} \small
\hat{Y}_{un}=f_{\theta}(X_{un})
\label{eq:caly}
\end{equation}


\newcommand{\argtopk}{\mathop{\mathrm{arg\,top}\,k}}

\begin{equation} \small
X_{sel}=\argtopk(SNR(\hat{Y}_{un}))
\label{eq:rank}
\end{equation}

\begin{equation} \small
SNR=\frac{\sum_{argmax(P)-\Delta_{F}}^{argmax(P)+\Delta_{F}}P_{i}}{\sum_{i=a}^{b}P_{i}}
\label{eq:SNR}
\end{equation}

\begin{equation} \small
k=R*N_{un}
\label{eq:K}
\end{equation}

\begin{equation} \small
R=m+n*e_j/e_{total}
\label{eq:ratio}
\end{equation}

where $\hat{Y}_{un}$ is the predicted pseudo-labels and ${X}_{un}$ is the unlabeled video clips. $a$ and $b$ are the low and high bandlimit which are set as 0.67 Hz and 3 Hz respectively corresponding with the typical resting heart rate of 40 BPM and 180 BPM. $P_{i}$ is the power for the $i$th frequency of the signal. $\Delta_{F}$ is the frequency difference to the peak frequency of $argmax(P)$. $\Delta_{F}$ is set as 0.1 Hz in this study. $X_{sel}$ is the selected unlabeled video clips. $k$ is the number of selected unlabeled video clips. $N_{un}$ is the total number of unlabeled video clips. $R$ is the ratio of selected unlabeled clips among all unlabeled labels. $e_j$ is the $j$-th epoch. $e_{total}$ is the number of the total epochs. $m$ and $n$ are hyperparameters that are set as 0.2 and 0.6 respectively in this study corresponding with the $R$ of 20\% and 80\%.

\subsection{Self-supervised consistency regularization}
\label{sec:consloss}

A self-supervised consistency regularization is proposed for quasi-periodic rPPG signals to extract the temporal features and prevent the model from overfitting. The consistency loss is inspired by FlexTeacher method \cite{zhang2021flexmatch} by assuming consistency between a strong augmentation and a weak augmentation of the original input. Since the rPPG signal is quasi-periodic, it possesses several unique characteristics. Some studies resample the clips to alter the predicted heart rate \cite{gideon2021way,li2023contactless}. In this study, we reverse the facial clips to obtain the strongly augmented clips $X_{s} \in \mathbb{R}^{T\times W \times H\times C}$. Reversing the facial clips will invert the pulse signal but the reversed pulse signal retains the same PSD distribution as the original signal. Besides, a minor temporal shift of the video clips will yield the same heart rate prediction. Hence, weakly augmented clips $X_{w} \in \mathbb{R}^{T\times W \times H\times C}$ are created by applying a small temporal shift of the original clips. Consequently, a consistency loss $L_{c}$  is developed to measure the agreement between the modified clips $x_{s}$ and $x_{w}$.

\begin{equation} \small
L_{c}=CE(PSD(f_{\theta}(x_{s})),PSD(f_{\theta}(x_{w})))
\label{eq:cons}
\end{equation}

\begin{figure}[t]
\centering
\includegraphics[scale=0.3]{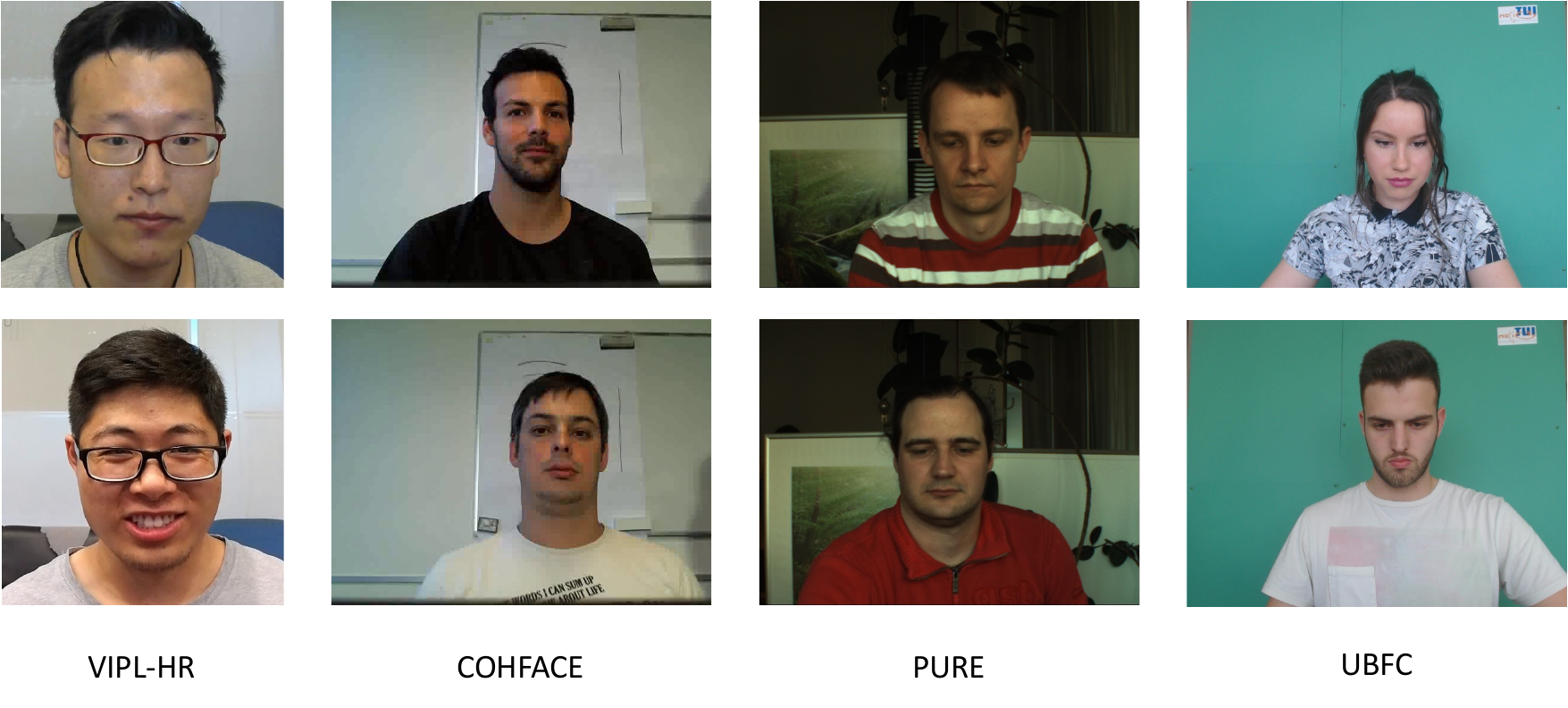}
  \caption{
  Illustration of training datasets: the four datasets have different races, skin tones, illumination conditions, backgrounds, and recording devices. There are domain gaps among various datasets. The numbers of subjects for these datasets are 107, 40, 10, and 42. The colored videos for each dataset are 2378, 160, 60, and 42, respectively. The subject's number and video number are relatively small, especially for the two most commonly used datasets of PURE and UBFC.
  }
\label{fig:datasets}
\end{figure}

\section{Experiment}
\label{sec:experiment}
The proposed Semi-rPPG method has been evaluated on four public datasets of VIPL-HR \cite{niu2019vipl}, PURE \cite{stricker2014non}, UBFC-rPPG \cite{bobbia2019unsupervised}, and COHFACE \cite{heusch2017reproducible} as illustrated in Fig.\ref{fig:datasets}. VIPL-HR and COHFACE are selected as intra-dataset evaluations due to their relatively large size so that more data can be used as unlabeled data. The cross-dataset evaluation is performed from UBFC to PURE using COHFACE as the unlabeled dataset. Besides, three traditional semi-supervised learning (SSL) methods are implemented on the same protocol and compared with the proposed Semi-rPPG.

\subsection{Datasets and performance metrics}
VIPL-HR~\cite{niu2019vipl} is a database with face videos of 107 subjects and the corresponding heart rate, SpO2, and BVP wave. The data were collected under 9 scenarios under various illumination and motion conditions with three camera devices. This database has 2,378 visible light videos (VIS) and 752 near-infrared (NIR) videos. We only use visible light videos for training.

PURE~\cite{stricker2014non} dataset consists of 10 subjects and each subject was recorded a 1-minute video under 6 scenarios. The videos were captured at a frame rate of 30 Hz with a cropped resolution of 640x480 pixels. The pulse rate wave and SpO2 readings with a sampling rate of 60 Hz are recorded in parallel with the videos.

UBFC-rPPG~\cite{bobbia2019unsupervised} dataset has 42 videos from 42 subjects. The video was captured at 30fps with a resolution of 640x480 in uncompressed RGB format. Ground-truth PPG waveform was obtained simultaneously. 

COHFACE~\cite{heusch2017reproducible} has 160 one-minute videos of 40 subjects. The heart rate and breathing rate of the recorded subjects are synchronized with the videos. The videos have been recorded at a resolution of 640x480 pixels and a frame rate of 20Hz. 

The experiment results are assessed based on heart rate measurement. The common metrics for rPPG learning of mean absolute error (MAE), root mean square error (RMSE), Pearson’s correlation coefficient (r), and standard deviation (SD) are employed to analyze the results.

\subsection{Implementation Details}
\label{sec:imp}

In this study,  an end-to-end 3D PhysNet is utilized where a sequence of facial frames is directly fed into the model for training. The face region of each frame is detected by MTCNN \cite{xiang2017joint}, then cropped and resized to $128*128$ pixels. The frequency of the pulse signals is resampled to match the frame rate of the facial clips. The model takes facial clips $X \in \mathbb{R}^{T\times W \times H\times C}$ as input and produces the pulse signal $y \in \mathbb{R}^{T}$ as output. $T$, $W$, $H$, and $C$ denote the frame number, width, height, and channels, respectively. The input and output have the same temporal length as the clips. The VIPL-HR dataset is divided into training (66 subjects), validation (20 subjects), and test datasets (21 subjects). For the COHFACE dataset, the first 32 subjects are used for training, and the last 8 subjects' data are used for testing.
The model is implemented with Pytorch. During supervised training, the batch size is set to 4, while for semi-supervised training, it is set to 2. The temporal length of each clip is set as 300 frames. We adopt Adam optimizer \cite{loshchilov2017decoupled} with learning rates of $1\times10^{-4}$ and $1\times10^{-5}$. 

\begin{table}[t]
\centering
\caption{Dataset split for VIPL-HR} 
\resizebox{0.46\textwidth}{!}
{
     \begin{tabular}{ c c c c c c }
      \toprule
      Protocol & Training & Training subjects & Training subject  & Validation & Test  \\ & & with labels & without labels & subjects& subjects
      \\
      \hline
      1 & Fully Supervised  & 66 & 0  & 21  & 20    \\
      2  & Partially Supervised & 13 & 0  & 21  & 20  \\
      3  & Semi-supervised  & 13 & 53 & 21  & 20 \\ 
      \hline
    \end{tabular}}
\label{tab:splVIPL}
\end{table}

\begin{table}[t]
\centering
\caption{Dataset split for COHFACE}
\resizebox{0.46\textwidth}{!}
{
    \begin{tabular}{ c c c c c }
      \toprule
      Protocols & Training & Training subjects & Training subjects  & Test  \\ & & with labels & without labels & subjects
      \\
      \hline
      1   & Fully Supervised & 32 & 0   & 8  \\
      2   & Partially Supervised & 6 & 0   & 8  \\
      3   & Semi-supervised & 6 & 26   & 8  \\ 
      \hline
    \end{tabular}}
\label{tab:splCOHFACE}
\end{table}

\subsection{Intra-dataset testing}
The proposed Semi-rPPG has been assessed using the VIPL-HR and COHFACE datasets for intra-dataset evaluation. The dataset split protocols are illustrated in Table~\ref{tab:splVIPL} and Table~\ref{tab:splCOHFACE}, while the experimental results are shown in Table~\ref{tab:intraVIPL} and Table~\ref{tab:IntraCOHFACE}. Protocol 1 and Protocol 2 are used for fully supervised training and partially supervised training as comparisons to semi-supervised learning with Protocol 3. The fully supervised training utilizes the entire training dataset with the loss function described in Eq.(\ref{eq:sup}). The partially supervised training only uses 20\% of the training data with the same loss as fully supervised training. Protocol 3 utilizes 20\% of the training data as labeled data while the remaining 80\% of training data is used as unlabeled data.

To compare with the proposed Semi-rPPG, three traditional semi-supervised learning (SSL) methods are implemented and compared under the same data split Protocol 3. A self-training method \cite{amini2022self} with the same SNR curriculum pseudo-labeling without consistency regularization is implemented. The other two consistency-based SSL methods of FixMatch \cite{sohn2020fixmatch} and FlexMatch \cite{zhang2021flexmatch} are implemented with the proposed consistency loss (Eq.(\ref{eq:cons})) in this study.

As depicted in Table~\ref{tab:intraVIPL}, the RMSE of partially supervised training is increased by 30.5\% compared with the fully supervised dataset. As partially supervised training used only 20\% labeled data, it demonstrates that the model is underfitting due to the limited amount of labeled information. By involving the unlabeled dataset, all four evaluation criteria of MAE, RMSE, r and SD are all improved for the three SSL methods compared with the baseline of partially supervised results. It proves the effectiveness of leveraging the unlabeled data for rPPG learning can improve the model performance when the labeled data is limited. In the proposed Semi-rPPG, the two novelties of curriculum pseudo-labeling and the rPPG consistency regularization are combined and the results outperform all the other SSL methods. The results are even comparable to the fully supervised results with RMSE difference of only 0.28 beats per minute (BPM).

\begin{table}[t]
\centering
\caption{Intra-dataset test results for VIPL-HR}
\resizebox{0.46\textwidth}{!}
{
    \begin{tabular}{ c c c c c c}
      \toprule
      Method & Dataset split & MAE & RMSE & r & SD  \\ & protocol & & & &  \\
      \hline
      Fully Supervised & 1    & 5.82    & 10.71   & 0.63  & 10.71     \\
      Partially Supervised & 2  & 7.97    & 13.98   & 0.42  & 13.92     \\
      \hline
      SSL-Self-training~\cite{amini2022self}& 3  & 7.10    & 13.14   & 0.49  & 13.12     \\
      SSL-FixMatch~\cite{sohn2020fixmatch} & 3   & 6.59  & 12.14   & 0.54  & 12.01     \\
      SSL-FlexMatch~\cite{zhang2021flexmatch} & 3  & 6.93   & 12.82   & 0.50  & 12.45     \\
      Semi-rPPG (Ours)  & 3  & \textbf{6.06}    & \textbf{10.99}   & \textbf{0.57}  & \textbf{10.98}     \\
      \hline
    \end{tabular}}
\label{tab:intraVIPL}
\end{table}

\begin{table}[t]
\centering
\caption{Intra-dataset test results for COHFACE}
\resizebox{0.46\textwidth}{!}
{
    \begin{tabular}{ c c c c c c}
      \toprule
      Method & Dataset split & MAE & RMSE & r & SD  \\ & protocol & & & &  \\
      \hline
      Fully Supervised & 1  & 1.91    & 3.70   & 0.95  & 3.68     \\
      Partially Supervised & 2  & 4.56    & 8.20   & 0.82  & 7.42     \\
      \hline
      SSL-Self Learning~\cite{amini2022self} & 3  & 6.66    & 10.05  & 0.69  & 8.83     \\
      SSL-FixMatch~\cite{sohn2020fixmatch} & 3   & 3.28  & 6.78   & 0.85  & 6.68     \\
      SSL-FlexMatch~\cite{zhang2021flexmatch} & 3  & 3.16   & 5.38   & 0.90  & 5.30     \\
      Semi-rPPG (Ours)  & 3  & \textbf{2.97}    & \textbf{5.11}   & \textbf{0.90}  & \textbf{5.11}     \\
      \hline
    \end{tabular}}
\label{tab:IntraCOHFACE}
\end{table}

The intra-dataset testing is also evaluated on COHFACE datasets. As COHFACE dataset is smaller than VIPL-HR, no validation dataset is utilized allowing more of the data to be allocated as unlabeled. A total of 32 subjects' data are used for fully supervised training. 20\% of the training data which is 6 subjects data are utilized for partially supervised as the baseline. Table~\ref{tab:IntraCOHFACE} reveals that the partially supervised training using 20\% labeled data has worse performance with an RMSE of 8.20 BPM compared to the fully supervised training which has an RMSE of 3.70 BPM. It suggested that the model may not work well with a small amount of labeled datasets for supervised training. The three SSL methods are implemented for the COHFACE dataset as well.
Self-learning has the worst performance on COHFACE dataset while FixMatch and FlexMatch both perform better than partially supervised learning. In contrast, the proposed Semi-rPPG achieves the best results among all SSL methods with a minimum RMSE of 5.11 BPM, which is 37.7\% lower than partially supervised training.

\subsection{Cross-dataset testing}
The proposed Semi-rPPG is also evaluated on a cross-dataset setting from UBFC-rPPG to PURE using COHFACE as unlabeled dataset. The baseline is established by leveraging all the UBFC-rPPG data as training data and all the PURE as the test dataset using supervised learning with loss in Eq.(\ref{eq:sup}). The proposed Semi-rPPG is compared with three SSL methods by using UBFC-rPPG as the labeled dataset and COHFACE as the unlabeled dataset and PURE as the test dataset.

As presented in Table~\ref{tab:cross}, our supervised method outperforms other supervised methods in cross-dataset evaluation from UBFC-rPPG to PURE as presented in a rPPG-toolbox study \cite{liu2024rppg-toolbox}. Incorporating COHFACE as an unlabeled dataset significantly enhances the performance of all SSL methods in cross-dataset learning from UBFC-rPPG to PURE although the datasets of COHFACE, UBFC, and PURE are from three different domains. This indicates that SSL methods have the capability of reducing the domain gap between UBFC-rPPG and PURE by leveraging another domain dataset of COHFACE for rPPG learning.  Among the four SSL methods evaluated, our proposed Semi-rPPG achieves the best performance with the lowest RMSE of 1.01 BPM.

\begin{table}[t]
\setlength\tabcolsep{3pt}
\centering
\caption{Cross-dataset evaluation from UBFC to PURE} 
\resizebox{0.46\textwidth}{!}
{
    \begin{tabular}{c| c c c c c c c}
      \hline
      Strategy& Method & Train & Test  & MAE & RMSE & r & SD  \\  & & dataset & dataset & & & &  \\
      \hline
      & Ours & UBFC-rPPG &PURE  & 2.35    & 7.49   & 0.95  & 7.29    \\
     Supervised &TS-CAN \cite{liu2020multi} & UBFC-rPPG &PURE  & 3.69   & -   & -  & -    \\
      &PhysFormer \cite{yu2022physformer} & UBFC-rPPG &PURE  & 5.54    & -   & -  & -    \\
      &DeepPhys \cite{chen2018deepphys} & UBFC-rPPG &PURE  & 5.47    & -   & -  & -    \\
      \hline
      &SSL-Self-training~\cite{amini2022self} & UBFC-rPPG+ & PURE & 0.91  & 2.24    & 0.95  & 2.22\\ & & COHFACE(un) & & & & &    \\
      
     Semi-supervised &SSL-FixMatch~\cite{sohn2020fixmatch} & UBFC-rPPG+ & PURE &  0.95  & 3.54    & 0.99  & 3.5\\  & & COHFACE(un) & & & & &      \\
      
      &SSL-FlexMatch~\cite{zhang2021flexmatch} & UBFC-rPPG+ & PURE & 2.17  & 9.16    & 0.92  & 9.14\\  & & COHFACE(un) & & & & &       \\

      &Semi-rPPG (Ours)  & UBFC-rPPG+ & PURE & \textbf{0.57}    & \textbf{1.01}   & \textbf{0.99}  & \textbf{0.99} \\ & &COHFACE(un) & & & & &     \\
      \hline
    \end{tabular}}
\label{tab:cross}
\end{table}

\subsection{Ablation study}

The ablation study is conducted to evaluate the proposed curriculum pseudo-labeling strategy, consistency loss, training model, and hyperparameters. In the curriculum pseudo-labeling strategy, a dynamically increasing ratio $R$ based on SNR for unlabeled data selection is adopted. To evaluate this strategy, comparisons are made with a series of fixed selection ratios and a dynamically decreasing ratio. Additionally, the irrelevant power ratio (IPR) is also compared as an alternative selection criterion. To assess the effectiveness of the proposed consistency loss in Eq.(\ref{eq:cons}), it is substituted with two other self-supervised losses \cite{sun2022contrast,gideon2021way} while the other settings are kept the same. The model of PhysFormer is trained under the same protocols and compared with the PhysNet model. Besides, the pre-training epoch and hyperparameter $\lambda$ are also evaluated in the study.

\begin{table}[t]
\setlength\tabcolsep{3pt}
\centering
\caption{Various sample ratio scenarios' evaluation results on VIPL-HR dataset} 
\resizebox{0.46\textwidth}{!}
{
    \begin{tabular}{ c c c c c c}
      \toprule
      Method & Dataset split & MAE & RMSE & r & SD  \\ & protocol & & & &  \\
      \hline
      Supervised & 2  & 7.97    & 13.98   & 0.42  & 13.92    \\
      \hline
      Semi-rPPG (R=90\%) & 3  & 8.03   & 13.37   & 0.41  & 13.23     \\
      Semi-rPPG (R=50\%) & 3  & 7.40   & 13.03   & 0.43  & 12.67     \\
      Semi-rPPG (R=10\%) & 3  & 7.50   & 13.64   & 0.48  & 13.56     \\
      Semi-rPPG (R=80\%\textasciitilde20\%) & 3  & 8.06   & 14.14   & 0.42  & 14.04     \\
      Ours (R=20\%\textasciitilde80\%)  & 3  & \textbf{6.06}    & \textbf{10.99}   & \textbf{0.57}  & \textbf{10.98}     \\
      \hline
    \end{tabular}}
\label{tab:abSNR}
\end{table}

\subsubsection{Ratio $R$ for pseudo-label selection}

The ratio $R$ will determine the ratio of selected unlabeled data among all the unlabeled data after ranking their SNR of pseudo-labels in semi-supervised training. A few fixed and dynamic ratio scenarios are evaluated on VIPL-HR dataset. The results are demonstrated in Table~\ref{tab:abSNR}. As shown, $R$ has a significant effect on the model. For fixed ratios, the model performs best when the $R=50\% $ where the RMSE is reduced by 6.7\% compared with the baseline of the supervised method where no unlabeled data is used. If the $R$ is very small ( $R=10\% $) or very large ($R=90\%$), the RMSEs are reduced by 2.4\% and 4.4\%, respectively. This can be explained that a higher ratio of unlabeled data introduces more noise into the model particularly when there is limited labeled data which makes the model more susceptible to noisy pseudo-labels. Conversely, with a very small fixed ratio, the benefits of the SSL method are not fully realized. Therefore, a moderate $R$ is normally effective when adopting a fixed ratio. Two dynamic ratio scenarios are compared: one is to increase the $R$ from 20\% to 80\% while the other decreases $R$ from 80\% to 20\%. The result indicates that increasing $R$ yields better performance with the lowest RMSE of 10.99 BPM which represents a 21.4\% reduction compared to the supervised RMSE of 13.98 BPM. As training progresses, the model becomes more robust and the generated pseudo-labels become more reliable, which further enhances model performance. Conversely, starting with a lower $R$ and then increasing it may lead to the bias of the model due to the inclusion of noisy pseudo-labels in the early training stages. Hence, our proposed increasing ratio is effective for unlabeled data selection.

\begin{table}[t]
\setlength\tabcolsep{3pt}
\centering
\caption{Comparison of SNR and IPR criterion on VIPL-HR}
\resizebox{0.46\textwidth}{!}
{
    \begin{tabular}{ c c c c c c}
      \toprule
      Method & Dataset split & MAE & RMSE & r & SD  \\ & protocol & & & &  \\
      \hline
      Semi-rPPG (IPR, R=20\%\textasciitilde 80\%) & 3  &10.59   & 16.86   & 0.25  & 16.10     \\
      Semi-rPPG (SNR, R=20\%\textasciitilde 80\%) & 3  & \textbf{6.06}    & \textbf{10.99}   & \textbf{0.57}  & \textbf{10.98}     \\
      \hline
    \end{tabular}}
\label{tab:comSNRIPR}
\end{table}

\begin{table}[t]
\setlength\tabcolsep{3pt}
  \begin{center}
    \caption{Self-supervised loss comparison on VIPL-HR} \label{tab:comconsis}
    \begin{tabular}{ c c c c c c}
      \toprule
      Methods & Dataset split & MAE & RMSE & r & SD  \\ & protocol & & & &  \\
      \hline
      Gideon2021 \cite{gideon2021way} & 3  & 6.47   & 12.02   & 0.53  & 12.01     \\
      Contrast-Phys \cite{sun2022contrast}  & 3  & 6.22   & 11.55   & 0.57  & 11.54     \\
      Semi-rPPG (Ours)   & 3  & \textbf{6.06}    & \textbf{10.99}   & \textbf{0.57}  & \textbf{10.98}     \\
      \hline
    \end{tabular}
  \end{center}
\end{table}

\begin{table}[t]
\centering
\caption{Model comparison of 3D PhysNet and PhysFormer}
\resizebox{0.46\textwidth}{!}
{
    \begin{tabular}{c| c c c c c c }
      \hline
      Strategy& Model & Dataset split & MAE & RMSE & r & SD  \\  & & protocol &  & & &  \\
      \hline
      Fully & PhysFormer \cite{yu2022physformer} & 1  & 6.37  & 11.35  & 0.54  &11.28 \\
      
     Supervised &PhysNet \cite{sun2022contrast} & 1   & \textbf{5.82}   & \textbf{10.71}   &\textbf{0.63}  & \textbf{10.71}     \\
     \hline
     
     Partially &PhysFormer \cite{yu2022physformer}  & 2 & 10.48    & 15.95   & 0.27 & 15.94   \\
     
      Supervised &PhysNet \cite{sun2022contrast}  & 2   & \textbf{7.97}    & \textbf{13.98}  & \textbf{0.42}  & \textbf{13.92}   \\
      \hline
      Semi-rPPG &PhysFormer \cite{yu2022physformer}  & 3 & 9.44   & 14.66   & 0.27  & 14.46    \\
      
      &PhysNet \cite{sun2022contrast}  & 3  & \textbf{6.06}    & \textbf{10.99}   & \textbf{0.57}  & \textbf{10.98}    \\

      \hline
    \end{tabular}
}
\label{tab:PhysFormer}
\end{table}

\begin{table}[t]
\setlength\tabcolsep{3pt}
  \begin{center}
    \caption{Pre-training epoch evaluation}\label{tab:epoch}
    \begin{tabular}{ c c c c c c}
      \toprule
      Method & Dataset split & MAE & RMSE & r & SD  \\ & protocol & & & &  \\
      \hline
      Semi-rPPG ($e_{pre}=30$)  & 3  & 7.69   & 13.93   & 0.42  & 13.92     \\
      Semi-rPPG ($e_{pre}=10$)  & 3  & 6.40   & 11.65   & 0.52  & 11.65     \\
      Semi-rPPG ($e_{pre}=1$)   & 3  & \textbf{6.06}    & \textbf{10.99}   & \textbf{0.57}  & \textbf{10.98}     \\
      \hline
    \end{tabular}
  \end{center}
\end{table}

\begin{table}[t]
\setlength\tabcolsep{3pt}
  \begin{center}
    \caption{Evaluation of hyperparameter $\lambda$}\label{tab:lambda}
    \begin{tabular}{ c c c c c c}
      \toprule
      Method & Dataset split & MAE & RMSE & r & SD  \\ & protocol & & & &  \\
      \hline
      Semi-rPPG ($\lambda=1 $)  & 3  & 7.81   & 13.82   & 0.46  & 13.82     \\
      Semi-rPPG ($\lambda=0.1 $)  & 3  & 6.44   & 11.60   & 0.57  & 11.56     \\
      Semi-rPPG ($\lambda=0.01 $)   & 3  & \textbf{6.06}    & \textbf{10.99}   & \textbf{0.57}  & \textbf{10.98}     \\
      \hline
    \end{tabular}
  \end{center}
\end{table}


\subsubsection{Pseudo-label selection criterion of SNR and IPR}
\begin{figure}[t]
\centering
\subfigure[SNR variations via training epochs]{
		\begin{minipage}[b]{0.4\textwidth}
			\includegraphics[width=1\textwidth]{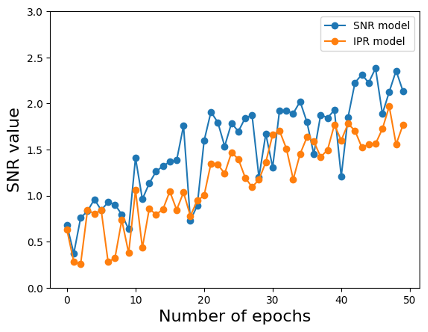}
		\end{minipage}
		\label{fig:snr}
	}\vspace{-2mm}
 \subfigure[IPR variations via training epochs]{
		\begin{minipage}[b]{0.4\textwidth}
			\includegraphics[width=1\textwidth]{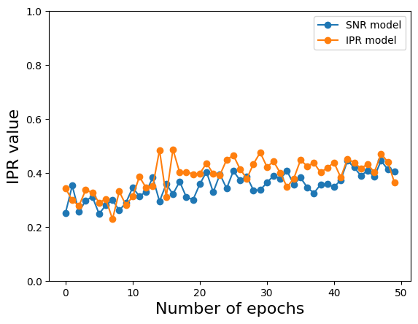}
		\end{minipage}
		\label{fig:ipr}
	}\vspace{-2mm}
\caption{The variations of SNR and IPR across training epochs}
\label{fig:snripr}
\end{figure}

The SNR is employed to rank the pseudo-labels as a signal quality criterion. Besides SNR, irrelevant power ratio (IPR) is another significant metric to assess the signal quality. IPR is defined as the ratio of irrelevant signal power to the total signal power as indicated in Eq.(\ref{eq:IPR}). The ablation experiment is conducted by replacing SNR with IPR while the other settings are kept the same. As presented in Table~\ref{tab:comSNRIPR}, the proposed Semi-rPPG model with SNR ranking criteria outperforms that of IPR criteria significantly in terms of MAE, RMSE, r, and SD. A sample from the validation dataset of VIPL-HR is selected to demonstrate how the SNR and IPR values evolve across the training epochs for both SNR and IPR models as depicted in Fig \ref{fig:snripr}.  It can be seen from Fig.\ref{fig:snr}, the SNR value increases significantly as the epoch grows for two models indicating the reduction of the noises with the model training. The SNR model generates higher SNR values for pseudo-labels compared with the IPR model which proves the superior performance of the SNR model. However, the IPR varies from 0.23-0.49 which is relatively stable in Fig.\ref{fig:ipr}. As the IPRs are below 0.5, it means the irrelevant signal power (frequencies beyond 0.67-3 Hz)) is smaller than the relevant signal power (heart rate frequencies within 0.67-3 Hz) all the time. Furthermore, the IPR value for SNR model is relatively lower than that of the IPR model, further supporting the superior performance of the SNR model. The IPR model imposes a relatively loose constraint, monitoring only the ratio of irrelevant to total signal power. In contrast, the SNR model imposes a stricter constraint by focusing on signals near the peak frequency. Therefore, the SNR model is more effective than the IPR model.

\begin{equation}
IPR=\frac{\sum_{i=-\infty}^{a}P_{i}+\sum_{i=b}^{\infty}P_{i}}{\sum_{i=-\infty}^{\infty}P_{i}}
\label{eq:IPR}
\end{equation}
where $P_{i}$ is the power for the $i$th frequency of the signal. $a$ and $b$  are the lower and upper bandlimit, respectively. In this study, $a$ and $b$ are set as 0.67Hz and 3Hz which corresponds to the heart rate of 40 BPM and 180 BPM.

\subsubsection{Consistency loss comparison}

To evaluate the effectiveness of the proposed consistency loss, another two self-supervised losses are assessed with the same curriculum pseudo-labeling method. Contrast-Phys~\cite{sun2022contrast} utilized a contrastive-based unsupervised method to extract physiological signals from facial videos. The model generated positive pairs from different spatiotemporal locations of the same video and negative pairs from two different videos. Gideon and Stent~\cite{gideon2021way} introduced a multi-view triplet loss for contrastive training by leveraging quasi-periodic signal characteristics. The positive samples wer taken from subset views of anchor samples. The negative samples were generated through a trilinear resampler. The comparison results are shown in Table~\ref{tab:comconsis}. It can be seen that our proposed consistency loss outperforms the other two self-supervised losses together with the same curriculum pseudo-labeling method. The RMSE of Semi-rPPG is reduced by 7.9\% and 4.8\% compared with self-supervised losses of Gideon2021 and Contrast-Phys, respectively. 

\subsubsection{Model comparison of 3D PhysNet and PhysFormer}
This study utilizes the 3D PhysNet \cite{sun2022contrast} as the training model.
The model of PhysFormer \cite{yu2022physformer} has been applied as the training model and compared with PhysNet in three data protocols as presented in Table \ref{tab:PhysFormer}. It can be seen that PhysNet outperforms the PhysFormer in both supervised and semi-supervised learning. Therefore, the PhysNet is adopted as the training model in this study. Besides, the Semi-rPPG method is also effective for PhysFormer achieving better results with a lower RMSE of 14.66 BPM compared to the partially supervised method which has an RMSE of 15.95 BPM when unlabeled data are used for training.

\subsubsection{Pre-training epoch evaluation}
In this study, the model is initially pre-trained with only labeled data during the first epoch. The pseudo-labels of the unlabeled data are generated from the pre-trained model. From the second epoch onward, the model is trained with original labeled data and pseudo-labeled data together. The pre-training epoch $e_{pre}$ is evaluated in Table \ref{tab:epoch}. As shown, the pre-training epoch has a significant effect on the model performance. The RMSE increases by 6.0\% and 26.75\% when the model is pre-trained with 10 epochs and 30 epochs respectively compared with 1 epoch in the pre-training stage. Given the limited amount of labeled data, the model is prone to overfitting with more pre-training epochs, which can lead to the generation of incorrect pseudo-labels and introduce bias into the training process.

\subsubsection{Evaluation of hyperparameter $\lambda$}
The hyperparameter $\lambda$ to trade off the two supervised losses of $L_{ce}$ and $L_{p}$ is evaluated in Table \ref{tab:lambda}. \textcolor{black}{As blood volume pulse signals are quasi-periodic signals, these signals have both frequency and phase characteristics. The cross entropy loss $L_{ce}$ can force the model to learn frequency features, while the negative Pearson correlation loss $L_{p}$ can force the model to have predicted output signals similar to ground-truth signals. This similarity includes both frequency and phase features. Therefore, both two losses are necessary for model learning. }

The $\lambda$ has a significant effect on the model. The model achieves the best performance when $\lambda=0.01$, outperforming the settings $\lambda=0.1$ and $\lambda=1$ by decreasing the RMSE of 5.26\% and 20.48\% respectively. A small $\lambda$ places more emphasis on the heart rate classification while giving less priority to the similarity between the predicted and ground-truth BVP signals.


\subsection{Discussion and Visualization}
\label{sec:Visualization}

\begin{figure}[t]
\centering
\includegraphics[scale=0.5]{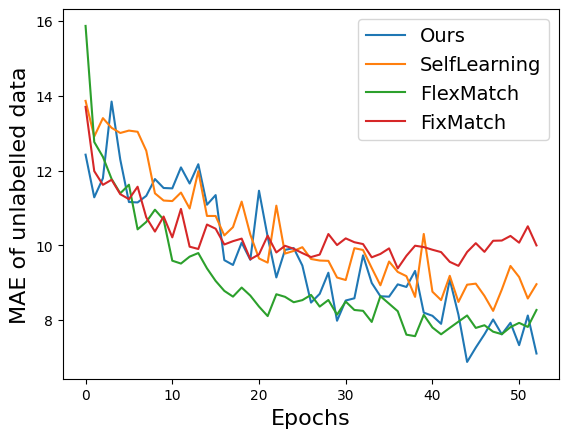}
  \caption{\small{
  MAE of unlabelled data with the training epochs
  }
  }
\label{fig:mae_unlabel}
\end{figure}

\begin{figure}[t]
\centering
\subfigure[Visualization of predicted blood volume pulse (BVP)]{
		\begin{minipage}[b]{0.4\textwidth}
			\includegraphics[width=1\textwidth]{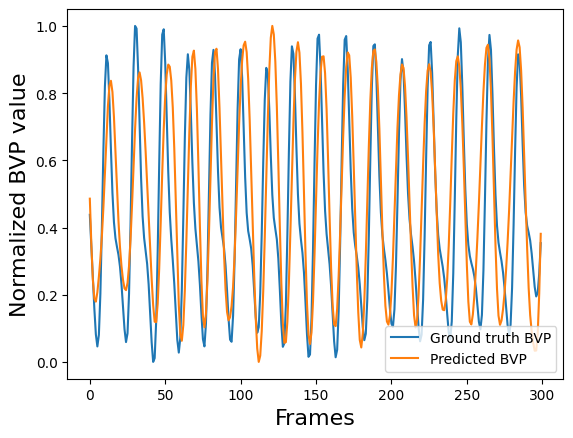}
		\end{minipage}
		\label{fig:bvp}
	}\vspace{-2mm}
 \subfigure[Visualization of predicted power density distribution (PSD)]{
		\begin{minipage}[b]{0.4\textwidth}
			\includegraphics[width=1\textwidth]{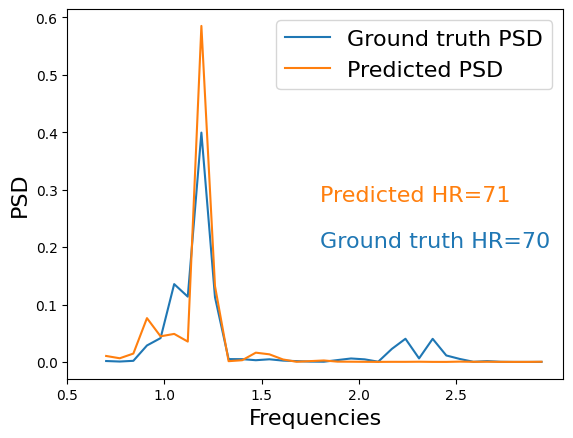}
		\end{minipage}
		\label{fig:psd}
	}\vspace{-2mm}
\caption{Visualization of the predicted and ground-truth BVP and PSD}
\label{fig:compare}
\end{figure}

\subsubsection{MAE of unlabelled data}
The mean absolute errors (MAEs) of the unlabelled data across four SSL methods in Table \ref{tab:intraVIPL} are plotted in Fig.~\ref{fig:mae_unlabel}. The MAEs of unlabeled data from the four methods decrease with the training of the model which proves the effectiveness of the four models. Although the MAEs of unlabeled data for FlexMatch methods initially drop faster than our proposed Semi-rPPG method, our approach achieves a lower MAE than FlexMatch after the model has converged. The minimum MAE for unlabelled data of Semi-rPPG is 6.06 BPM which is comparable to the test error of the fully supervised method whose MAE is 5.82 BPM. It demonstrates that the Semi-rPPG using 20\% labeled data and 80\% unlabeled data can achieve a comparable result with a supervised method using entire labeled data.

\subsubsection{Visualization of predicted BVP and PSD}
The ground-truth and predicted BVP and PSD are compared for a test sample of subject 102 from VIPL-HR dataset. The first 300 frames are adopted for evaluation. The BVP value is normalized from 0 to 1. As can be seen from Fig.~\ref{fig:bvp}, the predicted BVP aligns very well with the predicted BVP. By converting the BVP into PSD, the frequency distribution is depicted in Fig.~\ref{fig:psd}. The frequency of peak ground-truth PSD is 1.17 Hz and the frequency for peak predicted PSD is 1.19 Hz. As the heart rate is normally an integer, the corresponding heart rates are 70 and 71. From the comparisons of the BVP and PSD, the model not only learns well on the heart rate prediction but also learns the trending of the BVP signal. Further study can be explored to identify the abnormal BVP signal and corresponding heart rate diseases.

\begin{figure}[t]
\centering
\subfigure[A sample from VIPL-HR dataset]{
		\begin{minipage}[b]{0.2\textwidth}
			\includegraphics[width=0.8\textwidth]{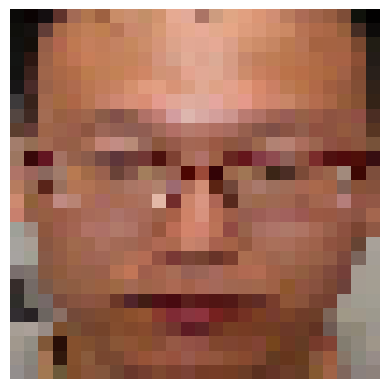}
		\end{minipage}
            \begin{minipage}[b]{0.24\textwidth}
			\includegraphics[width=0.8\textwidth]{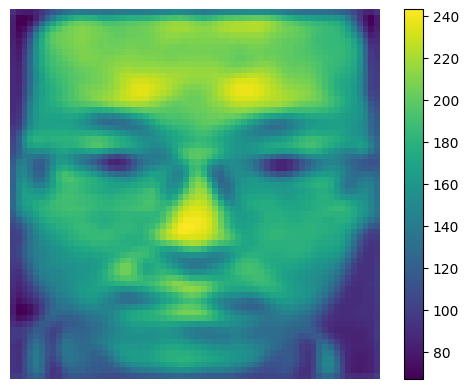}
		\end{minipage}
	}
 \subfigure[A sample from COHFACE dataset]{
		\begin{minipage}[b]{0.2\textwidth}
			\includegraphics[width=0.8\textwidth]{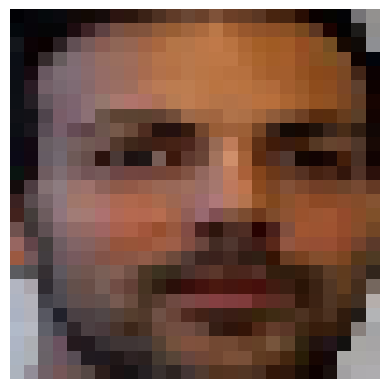}
		\end{minipage}
            \begin{minipage}[b]{0.24\textwidth}
			\includegraphics[width=0.8\textwidth]{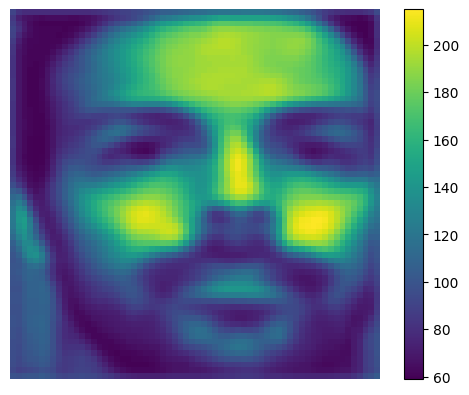}
		\end{minipage}
	}
\caption{Visualization of learned features. The brighter area indicates more attention from the model.}

\label{fig:colormap}
\end{figure}

\subsubsection{Color map of learned features}

The learned features in the loop layer of the model PhysNet are illustrated in Fig.~\ref{fig:colormap} using samples from VIPL-HR and COHFACE datasets. The brighter area indicates more attention from the deep learning model. It can be seen that the model mainly focuses on the skin of the forehead, cheeks, nose, and chin where rich PPG signals are contained.

\section{Conclusion}
\label{sec:conclusion}
In this study, we introduce a novel Semi-rPPG method that leverages a small amount of labeled facial videos and abundant unlabelled facial videos for rPPG deep learning. This approach incorporates an innovative curriculum pseudo-labeling strategy and a consistency regularization for quasi-periodic signals utilizing weak and strong augmentations for facial clips. The results reveal that the Semi-rPPG can significantly enhance the model performance in both intra-dataset and cross-dataset settings for rPPG learning. With 20\% of training data as labeled data and 80\% as unlabeled data, Semi-rPPG's performance is comparable to that of a fully supervised method that uses the entire training dataset as labeled data as demonstrated on the VIPL-HR dataset. The Semi-rPPG shows great potential to reduce the domain gap across different datasets as evidenced by its cross-dataset evaluation from UBFC-rPPG to the PURE dataset with COHFACE as the unlabeled dataset. Besides, the Semi-rPPG achieves the best results when compared with another three traditional semi-supervised methods. Ablation studies further evaluate the curriculum pseudo-labeling strategy and compare the consistency loss, showing that our proposed methods outperform other strategies.

\ifCLASSOPTIONcaptionsoff
  \newpage
\fi

\bibliographystyle{IEEEtran}
\bibliography{IEEEabrv,reference}


\end{document}